\documentclass[preprint,12pt]{elsarticle}

\usepackage{amssymb}
\usepackage{amsmath}
\usepackage{dsfont}
\usepackage{mathrsfs}
\usepackage{booktabs}
\usepackage{appendix}
\usepackage{graphicx}
\usepackage{subfig}
\usepackage[inkscape=latex]{svg}

\newcommand{\methodname}{LaDiMo}

\journal{arXiv}

\begin{document}

\begin{frontmatter}

\title{{\methodname}: Layer-wise Distillation Inspired MoEfier}
\author[sds]{Sungyoon Kim}
\ead{sy0319.kim@samsung.com}
\author[sds]{Youngjun Kim}
\ead{yj15.kim@samsung.com}
\author[sds]{Kihyo Moon}
\ead{kihyo.moon@samsung.com}
\author[sds]{Minsung Jang}
\ead{minsung.jang@samsung.com}
\address[sds]{Cloud Research Team, Samsung SDS}

\begin{abstract}
The advent of large language models has revolutionized natural language processing, but their increasing complexity has led to substantial training costs, resource demands, and environmental impacts. In response, sparse Mixture-of-Experts (MoE) models have emerged as a promising alternative to dense models. Since training MoE models from scratch can be prohibitively expensive, recent studies have explored leveraging knowledge from pre-trained non-MoE models. However, existing approaches have limitations, such as requiring significant hardware resources and data.
We propose a novel algorithm, {\methodname}, which efficiently converts a Transformer-based non-MoE model into a MoE model with minimal additional training cost. {\methodname} consists of two stages: layer-wise expert construction and routing policy decision. By harnessing the concept of Knowledge Distillation, we compress the model and rapidly recover its performance. Furthermore, we develop an adaptive router that optimizes inference efficiency by profiling the distribution of routing weights and determining a layer-wise policy that balances accuracy and latency. We demonstrate the effectiveness of our method by converting the LLaMA2-7B model to a MoE model using only 100K tokens, reducing activated parameters by over 20\% while keeping accuracy. Our approach offers a flexible and efficient solution for building and deploying MoE models.
\end{abstract}




\end{frontmatter}




\section{Introduction}
\label{sec:introduction}


The ascendance of Large Language Models (LLMs) has brought about a paradigm shift in the natural language processing (NLP) landscape, with their immense capacity to capture complex patterns and relationships in human language. However, this surge in model scale has also led to a concomitant increase in training costs, serving resources, and environmental footprints. As a response, the sparse Mixture-of-Experts (MoE) model has recently garnered significant attention as an alternative to dense models \citep{survey1}. The Feed-Forward Networks (FFNs) in Transformers are replaced by a set of experts, where only a subset of these experts are activated for each input token, thereby achieving computational efficiency \citep{moe}. For example, Mixtral-8x7B model \citep{mixtral}, a prominent large-scale model adopting the MoE architecture, reduces the number of active parameters by forwarding each input token only through the top 2 most relevant experts out of 8.

Since training an MoE model from scratch can be prohibitively expensive, recent studies have focused on leveraging knowledge from pre-trained non-MoE models. Moeficiation \citep{moefication} constructs experts based on neuron co-activation patterns, successfully converting ReLU-based T5 \citep{t5} and BERT \citep{bert} models into MoE models. However, this approach is limited because it is challenging to apply to recent state-of-the-art models employing different activation functions such as SwiGLU \citep{swiglu}. Meanwhile, \citet{llama-moe} proposed a method for transforming modern models like LLaMA \citep{llama} into MoE models but requires relatively large hardware resources and data for training.

Furthermore, in many MoE-based models, including Mixtral-8x7B, each token passed through the router is forwarded to a fixed number of experts. Nevertheless, since tokens exhibit varying levels of uncertainty, routing all tokens to the same number of experts at every layer can be inefficient \citep{wu2024, huang2024}. Researchers have recently investigated routing approaches that allow tokens to be routed to multiple experts dynamically to mitigate this issue, thereby enhancing performance and optimizing efficiency \citep{li2023, huang2024, zeng2024, lu2024}. While these existing methods necessitate some level of training or fine-tuning, to our knowledge, there has yet to be a proposal for adaptive routing strategies that do not require extra training.

To address the above issues, we introduce {\methodname}, a novel algorithm that construct an MoE model, namely \textit{MoEfy}, from a Transformer-based non-MoE model at a minimal additional training cost. {\methodname} consists of two stages: expert construction and routing policy decision. First, we construct an MoE model by leveraging the concept of Knowledge Distillation \citep{kd}, which utilizes the softmax output of a pre-trained model to train a more compact model. As shown in Figure \ref{fig:moefier}, through layer-wise distillation, where each expert learns to approximate the original layer's results, we achieve efficient model compression and rapid performance recovery. Subsequently, to optimize inference efficiency, we deploy an adaptive router. By profiling the distribution of routing weights computed by the router for input tokens, we determine a layer-wise policy that minimizes accuracy degradation and reduces inference latency. The contributions of this study can be summarized as follows:

\begin{itemize}
\item \textbf{Conversion to MoE with fast training, small data}: {\methodname} accelerates the transformation of a non-MoE model into a MoE model using Knowledge Distillation-based training. When applying our methodology to the LLaMA2-7B model, we successfully converted 12 layers into MoE layers using only 100K tokens, achieving an MMLU accuracy of over 97\% compared to the original model while reducing activated parameters by more than 20\%, resulting in significant computational cost savings.
\item \textbf{Layer-wise Model Optimization}: By selectively training and converting less influential layers of the original model into MoE layers, we preserve the original model's properties. After training, we set the layer-wise routing policy based on the behavior of each MoE layer, enabling additional throughput improvements.
\end{itemize}

\begin{figure}
  \centering
  \fontsize{5}{10}\selectfont
  \includegraphics[width=0.75\textwidth]{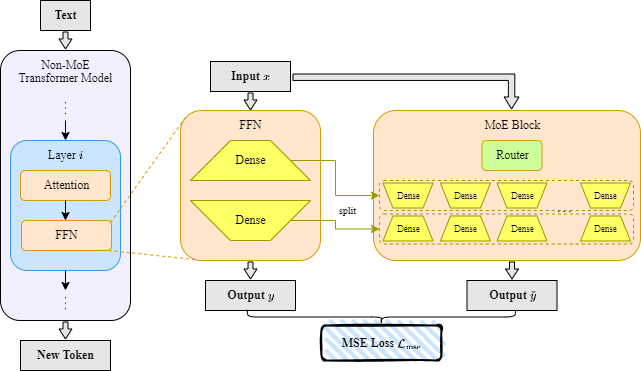}
  \caption{The main framework of Layer-wise Distillation Inspired MoEfier. The MoE block has its gating router and experts, which are FFNs whose weights are initialized by splitting the reference FFN's weight matrices. Input $x$ is obtained during inference tasks on a small text dataset. Those gathered inputs are used as the dataset for training the MoE block. Additionally, we have applied auxiliary loss and adaptive router, which will be explained in Sections \ref{subsec:auxiliary-loss} and \ref{subsec:adaptive-router}.}
  \label{fig:moefier}
\end{figure}


\section{Backgrounds}
\label{sec:backgrounds}
\subsection{Mixture-of-Experts}
\label{subsec:mixture-of-experts}

The concept of Mixture-of-Experts(MoE), wherein certain components of a model(i.e., experts) specialize in distinct tasks or knowledge domains, was initially introduced by \citet{firstmoe}. With the increasing scale of deep learning models, the Sparse MoE has been proposed in recent years, which aims to reduce computational costs by activating only a subset of experts \citep{moe}. Following this, the incorporation of MoE into Transformer-based large language models(LLMs) has yielded impressive performance gains \citep{gshard, switch}, leading to diverse research endeavors in this area. Notably, several industrial-scale LLMs incorporating MoE architectures, including Mixtral-8x7B \citep{mixtral}, DeepSeek-V2 \citep{deepseek2}, DBRX \citep{dbrx}, Grok-1 \citep{grok}, and Skywork-MoE \citep{skywork-moe}, have been released \citep{survey1}.

The most prevalent architecture for integrating MoE into Transformer-based models involves substituting the feed-forward network(FFN) within each Transformer block with a parallel $N$ FFNs $\{E_1, E_2, ..., E_N\}$, each constituting an individual expert, accompanied by a gating network, namely, a router. Specifically, for each input token $x$, the embedding vector is fed into the router $R$, which determines which experts to forward it to. A significant reduction in the computational cost of the FFN in dense models is achieved by activating a few experts.

\subsection{Knowledge Distillation}
\label{subsec:knowledge-distillation}

Knowledge distillation(KD) \citep{kd} is a prominent approach for compressing cumbersome pre-trained models into more compact and rapid models by leveraging their knowledge. The class probabilities generated by a teacher model with a large parameter set are utilized as soft targets, enabling a smaller student model to learn from these outputs. Employing high-entropy soft targets enables the distillation of more knowledge than exploiting hard targets.

As the complexity and scale of models increase, researchers have studied utilizing not only the output of the last layer but also intermediate representations from hidden layers \citep{romero2014}. Recently, studies on layer-wise distillation have been gaining traction, particularly with Transformer-based LLMs. For instance, \citet{sun2019} selectively leveraged hidden layers from a teacher model to fine-tune a smaller model for natural language processing tasks. Similarly, TinyBERT \citep{tiny-bert} adopted attention-based distillation and embedding layer-based distillation to reduce the computational requirements of the BERT model. Furthermore, \citet{chen2023} proposed a layer-wise distillation method that calculates the discrepancy between teacher model layers and student model layers using mean squared error(MSE) loss.


\section{Related Works}
\label{sec:related-works}
\subsection{From Dense to Sparse MoE Model}
\label{subsec:moefy}
When building an MoE model from the dense Transformer-based models, it is essential to determine which network components (e.g., FFNs, attention layers) to be replaced with experts and how (e.g., the total number of MoE layers, the number of experts per MoE layer) \citep{survey1}. While some researches have been conducted on converting attention layers into MoE structures \citep{zhang2022, jetmoe}, most studies have focused on converting FFNs. This is because FFNs account for a significant proportion of the overall FLOPs, and traditional ReLU-based models exhibit high activation sparsity in FFNs \citep{moefication, li2022, liu2023, zheng2024, pan2024}. Recent studies have explored converting models employing soft activation functions with relatively low activation sparsity (e.g., LLaMA's SwiGLU \citep{swiglu, llama}) into MoE models \citep{llama-moe, zheng2024}.

Determining the number of layers to be replaced with MoE layers, the number of experts per MoE layer, and the size of parameters of each expert is also crucial for designing MoE models. These hyperparameters directly impact the model's performance, including accuracy and system overheads such as memory requirements \citep{yun2024, krajewski2024, survey1}. For instance, MoE layers can replace either the entire model layers \citep{switch, mixtral, deepseek} or only specific layers \citep{gshard, stmoe}, and the position of the replaced layers can also affect performance (see Section \ref{subsec:layer-decision}). Our proposed methodology imposes no constraints on these configurations, balancing performance and execution efficiency.

Various approaches have been developed to construct MoE models by leveraging pre-trained weights from dense checkpoints. Sparse upcycling \citep{sparse-upcycling} constructs MoE layers by copying all parameters from the original dense model's FFNs to each expert. Building upon this concept, \citet{skywork-moe} empirically demonstrated that exploiting the original dense model's weights is more efficient when the budget for training a MoE model is limited. Moefication \citep{moefication} clusters and partitions intermediate FFNs based on co-activation patterns to construct experts. Inspired by this, \citet{zheng2024} proposed a method to learn non-ReLU activation models with an MoE structure efficiently. \citet{moebert} built experts based on FFN neurons according to their importance scores and performed layer-wise knowledge distillation from BERT models. Furthermore, \citet{llama-moe} suggested dividing original FFNs of SwiGLU-based models into multiple experts, although this approach requires relatively high costs for continued training. Considering these aspects, we propose a methodology that constructs inference-efficient sparse MoE models through layer-wise distillation while keeping training costs low.

\subsection{Expert Choice Strategies}
\label{subsec:expert-choice}
The efficiency of MoE models relies on the activation of only a subset of experts, and the expert selection strategy has a significant impact on model performance. Generally, for a given input $i$-th $x_i$, the output of the router $R$ can be represented as follows:
\begin{equation} \label{eq:router}
R(x_i) = (r_{i1},\ldots, r_{iN})
\end{equation}
where $r_{ij}$ denotes the probability of assigning the $i$-th token to expert $j$, also known as the routing weight \citep{moe}. Conventionally, a static value $k$ is set to be smaller than $N$, and the top-$k$ experts with the highest routing weights are selected \citep{moe, gshard, switch, skywork-moe}, such as in the case of the Mixtral-8x7B model where $k=2$ \citep{mixtral}.

Intuitively, if the maximum routing weight is sufficiently large, selecting only a single expert may have a negligible impact on the result while reducing computational costs. Conversely, if the distribution of routing weights is uniform, choosing multiple experts may benefit model accuracy (See Section \ref{sec:appendix:adaptive-router}). This implies that each token does not necessarily need to be forwarded to an equal number of experts, and recent studies have proposed dynamic routing strategies. \citet{li2023} optimized the training process by routing to either the top-$1$ or top-$2$ experts based on the weight difference between the highest expert and the second-highest one. \citet{wu2024} trained all experts during fine-tuning for tokens with uniform routing weight distributions and maintained the conventional approach at inference time, thereby improving model performance without incurring additional computational costs. \citet{huang2024} proposed a method that selects $n$ experts until the sum of their routing weights exceeds a certain threshold. Similarly, \citet{lu2024} chose only the top expert if the ratio of the highest to the second-highest weight exceeded a certain threshold. 

Besides optimizing routing strategies, efforts have been made to improve experts' architecture. \citet{li2023} achieved the top-all effect by merging experts according to their routing weight ratios, leading to improved model accuracy. \citet{zeng2024} boosted efficiency by introducing a FLOP-free null expert set and increasing the top-$k$. Yet, most existing studies suffer from the limitation of requiring additional training, whereas our proposed method determines the layer-wise routing policy in a training-free manner.


\section{Methodology}
\label{sec:methodology}

{\methodname} is focused on mimicking a given non-MoE model by constructing an MoE model and training to approximate the output with a limited text dataset and time. With such limited resources, training the whole layers with their FFNs replaced with MoE blocks as in previous approaches might degenerate due to the underfitted result \citep{moefication, llama-moe, sparse-upcycling}, which motivated us to substitute FFNs to MoE blocks for only some layers. To do so, how many layers to be selected and which layers to be selected should be considered. As the number of layers chosen increases, the FLOPs decrease, which gives a better throughput, while the accuracy also goes lower under restricted training resources (See \ref{subsec:layer-decision}). This trade-off needs to be treated carefully. By training some MoE blocks independently and assembling them, one can find an optimal composition of original layers from the reference model and newly trained layers.

\subsection{Continued Pre-training}
\label{subsec:continued-pre-training}

Given a layer from the transformer-based non-MoE model, the FFN in the layer consists of two dense projections. Some models like LLaMA use SwiGLU \citep{swiglu} as their activation function, whose FFN contains three dense projections. The FFN transforms an input $x \in \mathbb{R}^{d_h}$ into
\begin{equation}\label{eq:ffn}
    \left(\mathbf{x}W_u \odot \text{Swish}(\mathbf{x}W_g)\right)W_d,
\end{equation}
where $W_u \in \mathbb{R}^{d_h \times d_i}, W_g \in \mathbb{R}^{d_h \times d_i}, W_d \in \mathbb{R}^{d_i \times d_h}$ denote the up, gate, and down projection weights respectively, $d_h, d_i$ denote the dimension of hidden and intermediate state vector respectively, and $\odot$ denotes element-wise product.

To construct an MoE block, {\methodname} starts with splitting those weight matrices into $N$ submatrices where $N$ is the number of experts. The resulted submatrices are $W_u^{(i)} \in \mathbb{R}^{d_h \times d_i^{\prime}}, W_g^{(i)} \in \mathbb{R}^{d_h \times d_i^{\prime}}, W_d^{(i)} \in \mathbb{R}^{d_i^{\prime} \times d_h}$ with $d_i^{\prime} = \frac{d_i}{N}$ and $1 \leq i \leq N$, whose selected indices set is $I_i = \{(i-1)d_i^{\prime} + 1, (i-1)d_i^{\prime} + 2, \cdots, id_i^{\prime} \}$. The $i$-th expert is namely an FFN which transforms an input $x \in \mathbb{R}^{d_h}$ into
\begin{equation}\label{eq:ffn-moe}
    \left(\mathbf{x}W_u^{(i)} \odot \text{Swish}(\mathbf{x}W_g^{(i)})\right)W_d^{(i)}.
\end{equation}

Starting from the initial values from a pre-trained model's weight recovers the performance rapidly as in most similar approaches \citep{sparse-upcycling, solar, skywork-moe}. We verified that this approach boosts the training, as shown in Figure \ref{fig:continued-pretraining}.

\begin{figure}
    \centering
    \subfloat[4K steps]{%
        \includegraphics[width=0.5\textwidth]{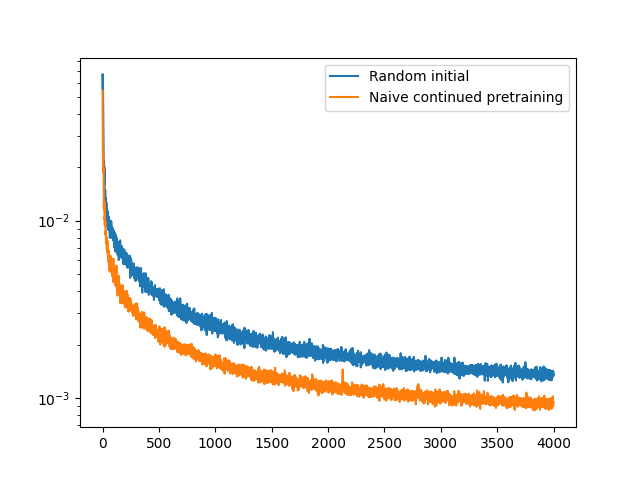}%
        \label{fig:continued-pretraining-4k}%
        }%
    \hfill%
    \subfloat[400K steps]{%
        \includegraphics[width=0.5\textwidth]{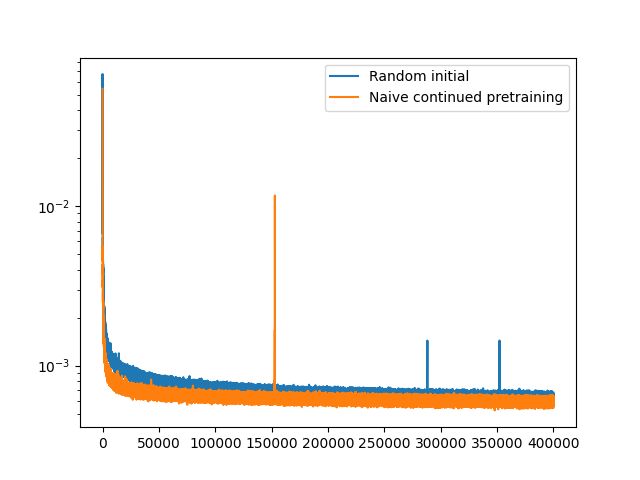}%
        \label{fig:continued-pretraining-400k}%
        }%
    \caption{Continued pre-training gives a smaller loss than starting from random initial weights. The experiment was conducted under NVIDIA A100 single GPU with the Chatbot Instruction Prompts dataset \citep{chatbot} and the LLaMA2-7B model.}
    \label{fig:continued-pretraining}
\end{figure}

\subsection{Layer-wise Distillation}
\label{subsec:layer-wise-distillation}

{\methodname} trains the MoE block to mimic an FFN as in layer-wise distillation methods. Unlike the original knowledge distillation \citep{kd}, which compares the last layer's output state, layer-wise variations train the hidden states at each layer, which is proven to improve the generalization performance \citep{sun2019, tiny-bert, chen2023}. Inspired by this approach, {\methodname} trains the MoE block by setting the loss function as
\begin{equation}\label{eq:mse-loss}
    \mathcal{L}_{\text{mse}} = \text{MSE}(\tilde{f}(\mathbf{x}), f(\mathbf{x}))
\end{equation}
provided we consider the FFN and MoE block as functions $f$ and $\tilde{f}$ respectively. Here $\text{MSE}(\cdot,\cdot)$ is the mean-squared error of two vectors.

In this scheme, the training data should be composed of inputs for the FFN, namely hidden states. The hidden tensors can be gathered from a sampled text dataset during inference tasks.

\subsection{Auxiliary Loss}
\label{subsec:auxiliary-loss}

In general, sparse computation using routing functions has a common issue of load imbalance among experts. The imbalance makes only a few experts to be used, which results in poor performance due to the limited parameters activated \citep{gshard}. To mitigate this issue, \citet{moe} suggested adding auxiliary losses to penalize the imbalance and encourage uniform routing. Some variations, such as a more straightforward form from Switch transformer \citep{switch}, are also used.

\begin{figure}
    \centering
    {\includegraphics[width=0.5\textwidth]{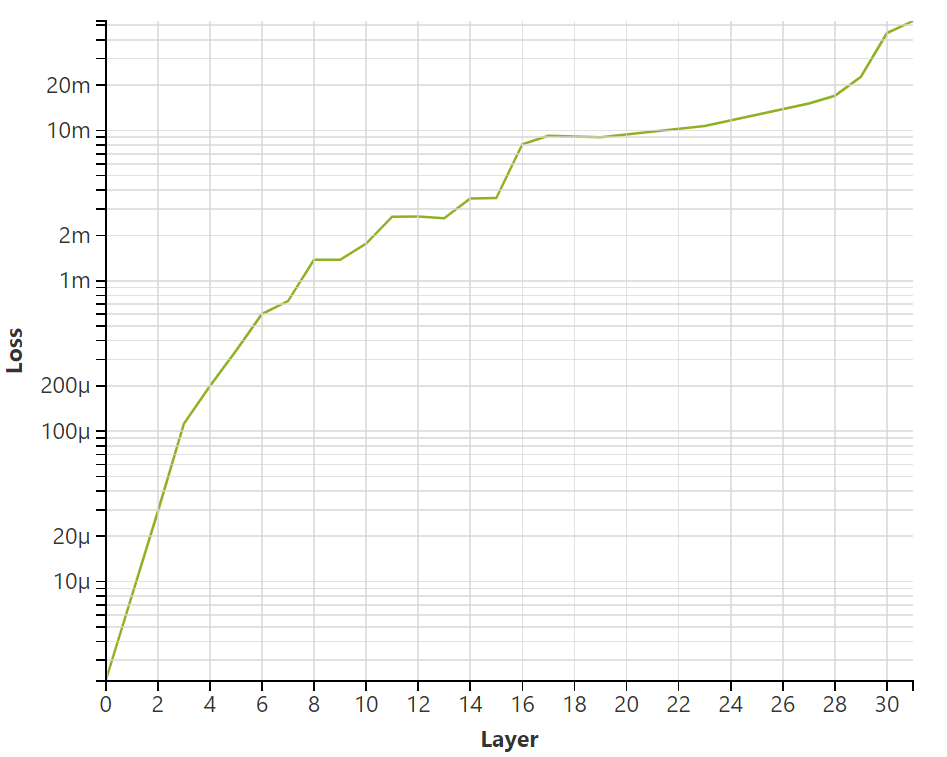}}
    \caption{Training loss $\mathcal{L}_{\text{mse}}$ for each layer's MoEfier.}
    \label{fig:layer-loss}
\end{figure}

The auxiliary loss of a given MoE layer can be written as
\begin{equation}\label{eq:aux-loss}
    \mathcal{L}_{\text{aux}} = \sum_{i=1}^N f_i P_i,
\end{equation}
where $f_i$ is the fraction of tokens out of the current batch dispatched to the $i$-th expert $i$, and $P_i$ is the fraction of the router probability given to the $i$-th expert. Switch Transformer combined two types of loss with an adjustment hyper-parameter $\alpha$ as follows:
\begin{equation}\label{eq:loss-sum}
    \mathcal{L_{\text{tot}}} = \mathcal{L}_{\text{mse}} + \alpha \mathcal{L}_{\text{aux}}.
\end{equation}
However, the coefficient $\alpha$ is not necessarily identical over layers \citep{skywork-moe}, regarding that the order of $\mathcal{L}_{\text{mse}}$ gets larger as the position of the layer varies from top to bottom, as shown in Figure \ref{fig:layer-loss}. Based on this phenomenon, we modified the loss scheme to
\begin{equation}\label{eq:loss-sum-modified}
    \mathcal{L_{\text{tot}}} = \mathcal{L}_{\text{mse}} + \alpha \lVert \mathcal{L}_{\text{mse}} \rVert \mathcal{L}_{\text{aux}}
\end{equation}
so that the scale of coefficient of $\mathcal{L}_{\text{aux}}$ be adaptively adjusted to keep the balance of incorporation of the two losses.

\subsection{Adaptive Router}
\label{subsec:adaptive-router}

Most MoE models are implemented with various versions of the top-$k$ router, initially proposed by \citet{moe}. Recently, some researchers have focused on adaptive routing, where the number of experts to be activated differ layer-by-layer, token-by-token, or both \citep{li2023, huang2024, guo2024, zeng2024}. Most adaptive router approaches need training or fine-tuning for their newly designed router, but it might not be affordable under limited resources. {\methodname} adaptively and dynamically route experts in a training-free way, mitigating the lack of training data. For a more detailed explanation of this approach, refer to Section \ref{sec:appendix:adaptive-router}.



\section{Experiments}
\label{sec:experiments}
We performed a series of experiments under 8 NVIDIA A100 GPUs. We have used DeepSpeed \citep{deepspeed} to train MoE blocks and vLLM \citep{vllm} to benchmark the inference latency of resulted model.

\subsection{Dataset}
\label{subsec:training-dataset}
To train our LaDiMo model, we needed to prepare the training dataset. Since the inputs of both FFN $f$ and MoE block $\tilde{f}$, defined in Section \ref{subsec:layer-wise-distillation}, should be hidden states, it was required to gather such inputs for each FFN selected to be transformed into a MoE block. We leveraged the Chatbot Instruction Prompts dataset \citep{chatbot} to obtain about 100K input vectors for each layer. For each training of the MoE block, we used these vectors as a training dataset with batch size 32 and 1M steps. Training for a single layer took about 5 hours.

\subsection{Layer Decision}
\label{subsec:layer-decision}
\begin{figure}
    \centering
    \subfloat[Throughput]{%
        \includegraphics[width=0.45\textwidth]{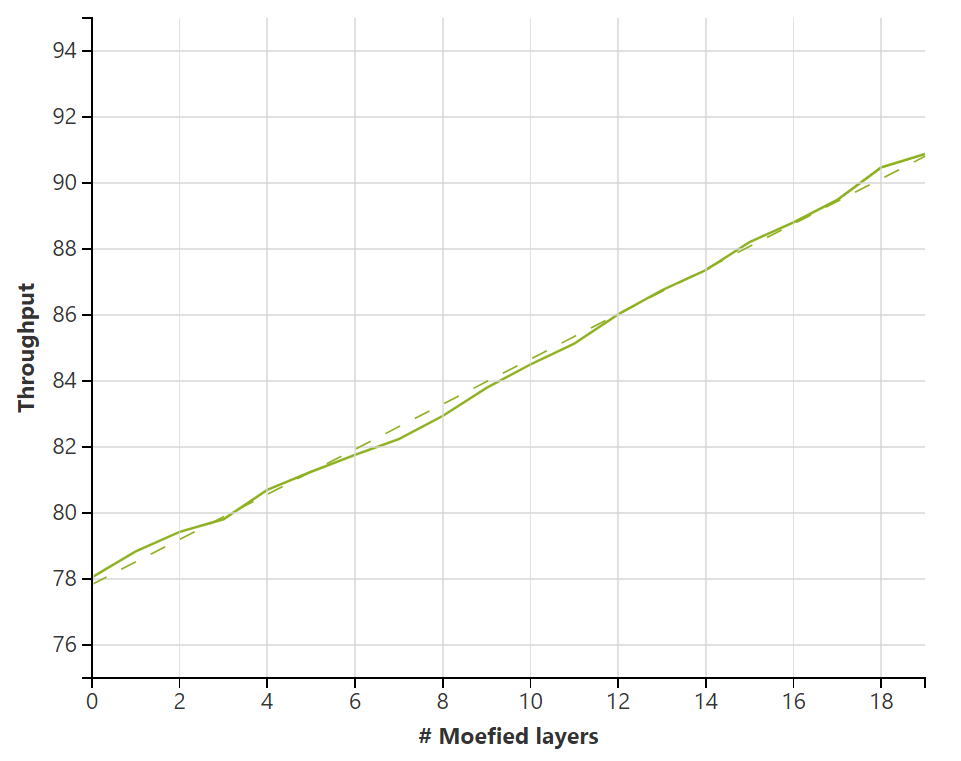}%
        \label{fig:moefied-layers-throughput}%
        }%
    \hfill%
    \subfloat[Accuracy]{%
        \includegraphics[width=0.45\textwidth]{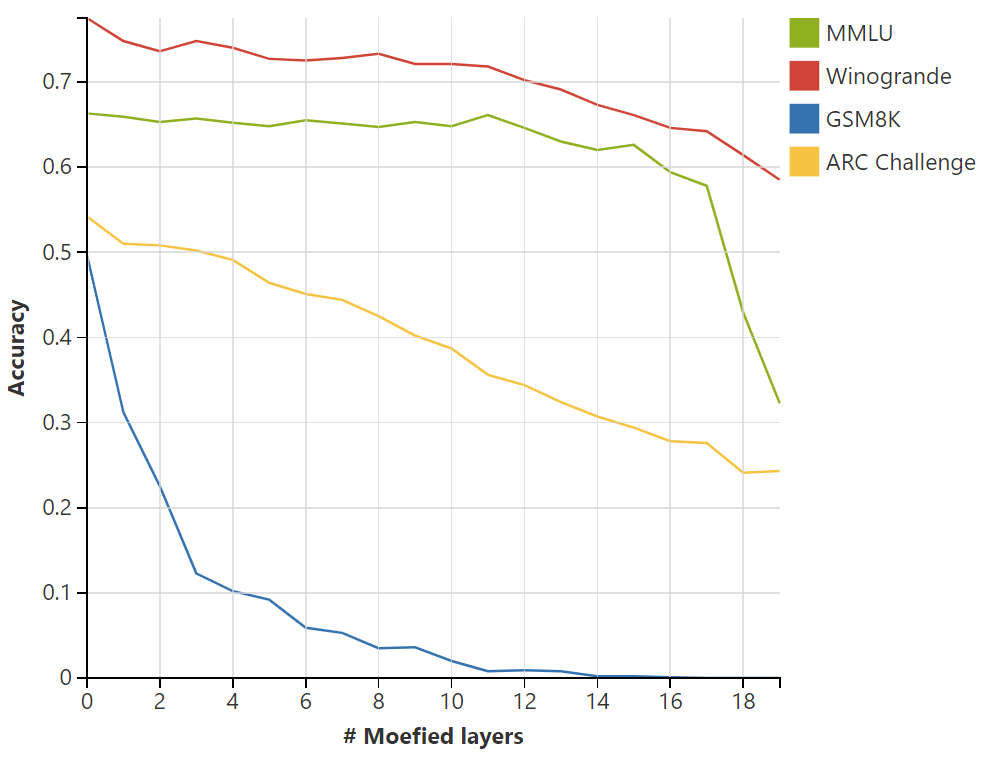}%
        \label{fig:moefied-layers-accuracy}%
        }%
    \caption{Effects of changes in the number of MoEfied layers into throughputs and accuracies for LLaMA-2 7B model.}
    \label{fig:moefied-layers-throughput-accuracy}
\end{figure}

\begin{figure}
    \centering
    {\includegraphics[width=0.7\textwidth]{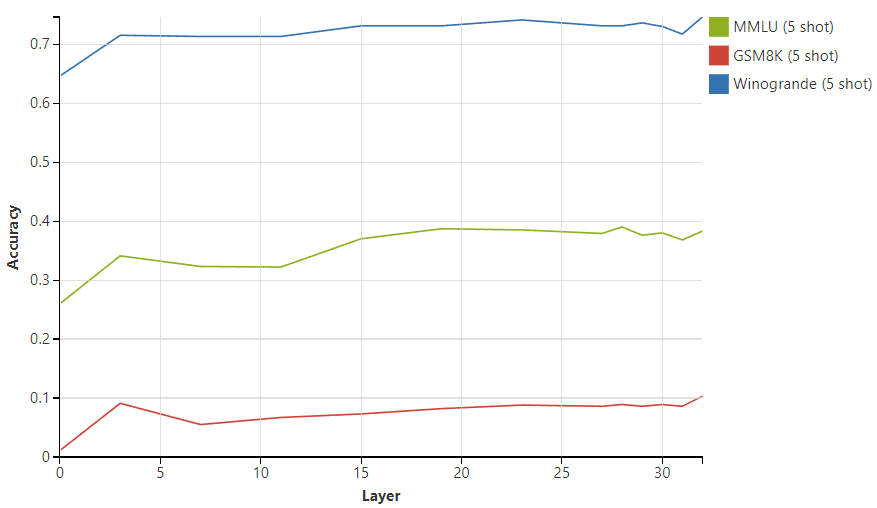}}    
    \caption{Relation between the accuracies and the MoEfied single layer. The $x$ axis refers to the layer index from 0 to 31, and additionally, the vanilla LLaMA-2 7B model's accuracies are plotted at $x=32$.}
    \label{fig:layer-accuracy}
\end{figure}

We observed that as the number of layers with its FFN replaced with trained MoE block increases, so does the throughput of the assembled partially MoEfied model, while the accuracy decreases (See Figure \ref{fig:moefied-layers-throughput-accuracy}). We also observed that if we replace the FFN from a single layer with a trained MoE block, the negative effect on its accuracy declines as the layer index goes to the end, as shown in Figure \ref{fig:layer-accuracy}. Regarding these observations, we performed experiments by replacing the bottom-most layers with MoE blocks, varying the number of the chosen layers.

\subsection{Results}
\label{subsec:results}
\begin{figure}
    \centering
    {\includegraphics[width=0.5\textwidth]{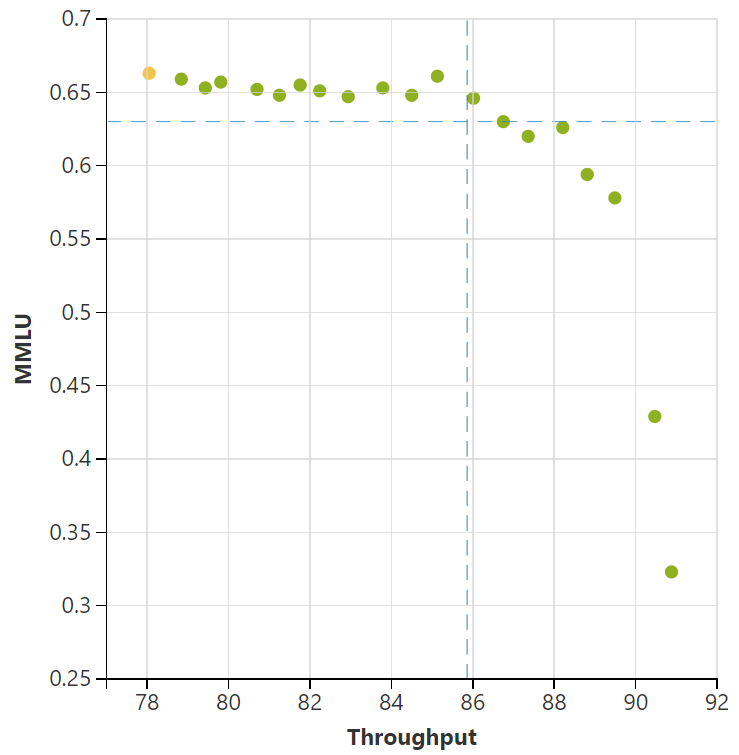}}    
    \caption{The MMLU accuracy and throughput for partially MoEfied LLaMA-2 7B models with various number of MoEfied layers. The yellow point indicates the original vanilla model, the blue horizontal dashed line marks 95\% of the vanilla model's MMLU score, and the blue vertical dashed line marks 110\% of the vanilla model's throughput.}
    \label{fig:throughput-mmlu}
\end{figure}

Figure \ref{fig:throughput-mmlu} shows the relations between throughput and MMLU accuracy with five shots for partially MoEfied models whose bottom-most $m$ layers MoEfied, where $0 \leq m < 20$. One can choose a proper model based on the trade-off between text quality and throughput. For instance, the partially MoEfied model with the last 12 layers MoEfied runs inference with its throughput enhanced 10\% while keeping 97\% MMLU accuracy of the LLaMA-2 7B model. This model has 6.7B parameters, almost the same as the original model, but the activated parameter size counts to 5.5B.

\subsection{Changing Dataset}
\label{subsec:changing-dataset}
\begin{figure}
    \centering
    {\includegraphics[width=0.5\textwidth]{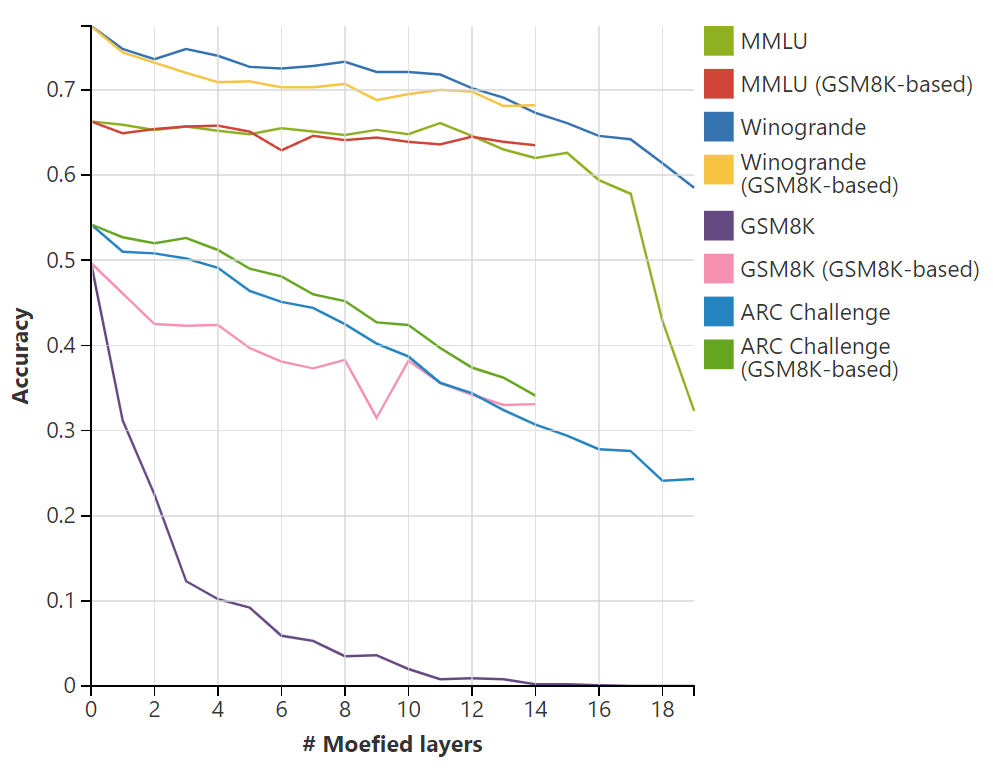}}
    \caption{Comparison of accuracy scores between training with a general text dataset and GSM8K.}
    \label{fig:moefied-layers-accuracy-gsm8k}
\end{figure}

Unlike other evaluation metrics using log-likelihood, GSM8K sharply declines as the number of MoEfied layers increases. This implies that a partially MoEfied model with MoE blocks trained with a limited amount of text dataset recovers its generation capability in a general sense but still lacks the ability to generate in a specific field. However, training MoE blocks with the GSM8K dataset \citep{gsm8k} can dramatically recover the score while keeping other evaluation scores. Thus, one can recover the text quality on particular fields on demand. This tendency is illustrated in Figure \ref{fig:moefied-layers-accuracy-gsm8k}.

\section{Conclusion}
\label{sec:conclusion}

In summary, we have successfully trained a couple of MoE blocks to mimic FFNs of a given non-MoE model and converted LLaMA2-7B model into partially MoEfied 5.5B model using only 100K tokens. The composition of MoE layers and original non-MoE layers can be customized to control the trade-off between throughput and accuracy. Still, some limitations remain, and further work can be extended.
\begin{itemize}
    \item We have selected the last layers to be MoEfied based on our observation. However, the process can be accomplished more carefully regarding the importance of the layer, namely the sensitivity of each layer to the output. A methodology that adaptively adjusts the configuration of each MoEfied layer, including the number and size of experts, can also be considered. Generalization of the adaptive router with layer policies top-$k$ with not only $k<4$ but also $k \geq 4$ would be accompanied along with this extension.
    \item We have performed experiments on the small size of datasets. However, extensive studies with more datasets would be helpful to understand how our approach can be effective in various situations.
\end{itemize}


\begin{appendices}
\section{Adaptive Router}
\label{sec:appendix:adaptive-router}

\begin{figure}
    \centering
    {\includegraphics[width=0.5\textwidth]{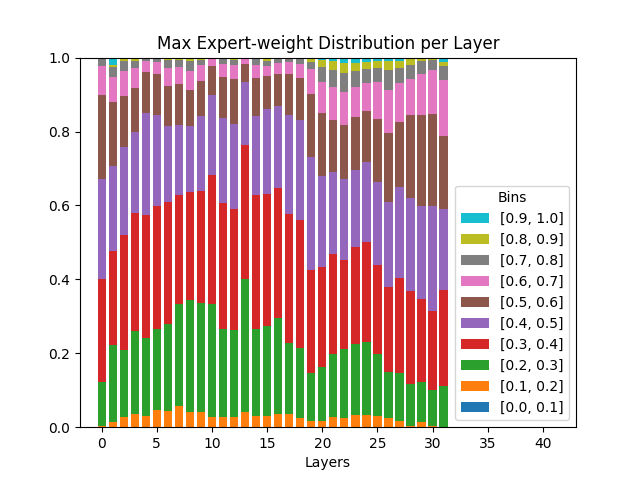}}
    \caption{Maximal routing weights distribution per layer, gathered during inference tasks.}
    \label{fig:max-expert-weight-distribution-per-layer}
\end{figure}

We assumed that (1) if a router assigns a large logit to a single expert, then dispatching the second-large expert becomes unnecessary, and (2) if experts are assigned evenly distributed logits, then it may need to consider more than two experts. To decide the case, we use the maximal routing weight over the experts as its standard since this value would be bigger in the former case and smaller in the latter case. Figure \ref{fig:max-expert-weight-distribution-per-layer} shows the distribution of the maximal routing weights differ layer-by-layer, which suggests that the decision of layer policy based on this value works.

\begin{figure}
    \centering
    {\includegraphics[width=1.0\textwidth]{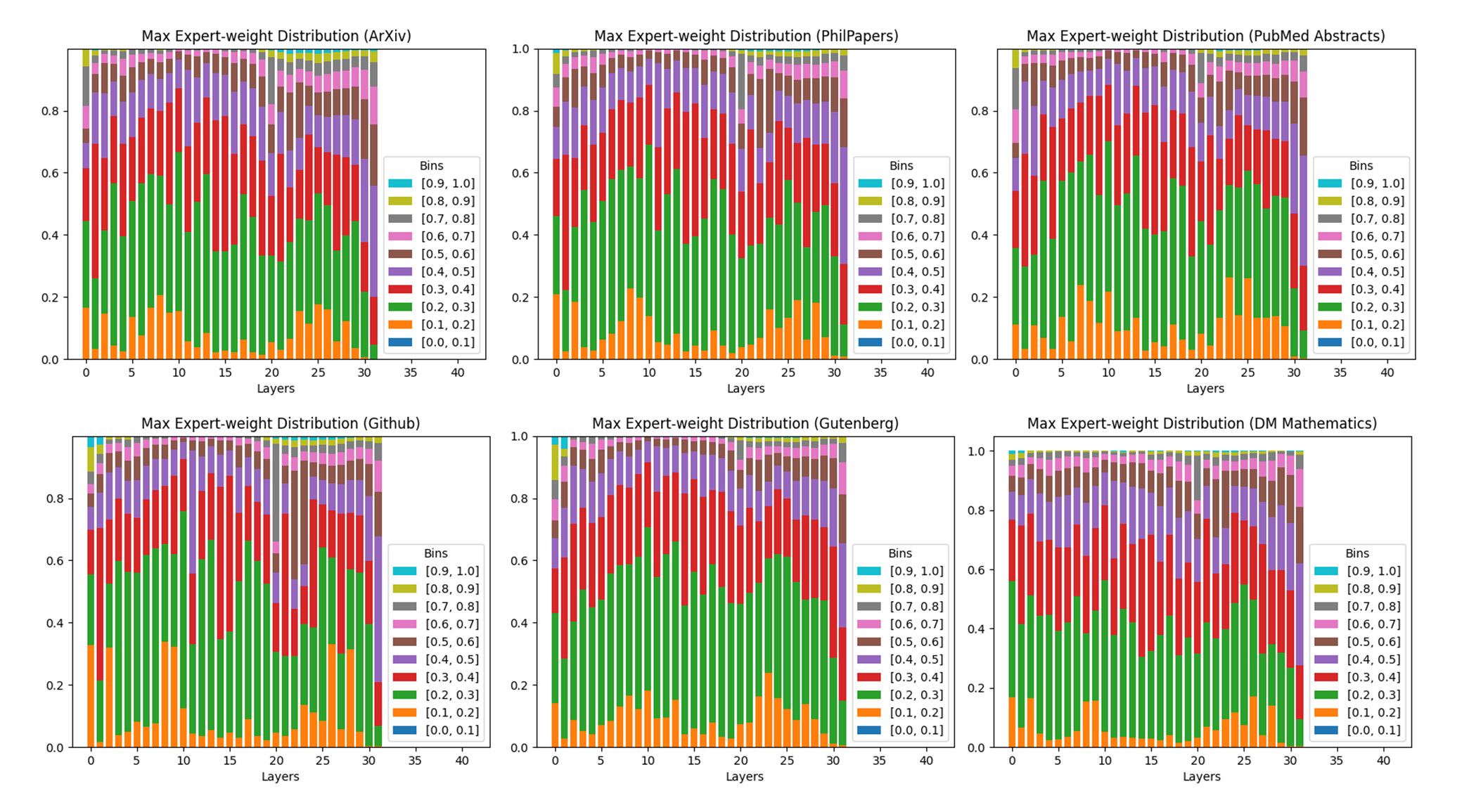}}
    \caption{Maximal routing weights distribution for various PILE datasets \citep{pile}. Datasets with similar text styles, e.g., ArXiv and PhilPapers, show a similar distribution. However, Github and DM Mathematics show different results.}
    \label{fig:max-expert-weight-distribution-datasets}
\end{figure}

As in Section \ref{subsec:layer-wise-distillation}, the maximal routing weight for each layer and each token can be accumulated during inference tasks on a sampled text dataset. We use these profiled weights to decide each layer's top-$k$ policy. This decision is based on our observation that while serving the LLM model, users' input prompts would form a word pool with its characteristic distribution of maximal routing weights. Refer to Figure \ref{fig:max-expert-weight-distribution-datasets} for a more detailed explanation.

\begin{figure}
  \centering
  \fontsize{5}{10}\selectfont
  \includegraphics[width=0.75\textwidth]{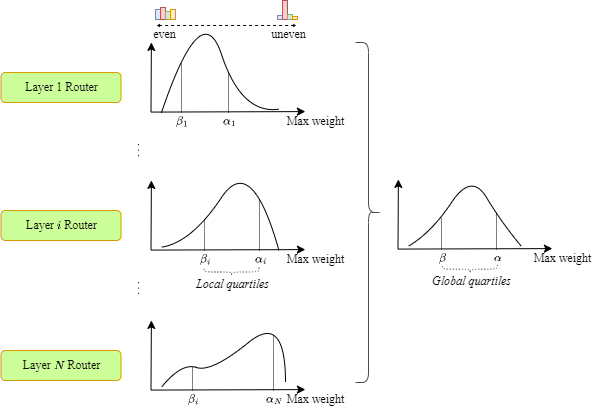}
  \caption{How the global quartiles $\alpha, \beta$ and local quartiles $\alpha_i, \beta_i$ are decided.}
  \label{fig:adaptive-router}
\end{figure}

\begin{enumerate}
    \item Given the hyper-parameter $p_u$ and $p_e$, which are set to be the ratio of extremely uneven/even routing weight distributions respectively, find two global quantiles $\alpha, \beta \in [0, 1]$ such that the probability $\underset{i \in \mathscr{L}}{\mathds{P}}(w_m \geq \alpha)$ of the maximal routing weight $w_m$ being greater than or equal to $\alpha$ over all layers $i \in \mathscr{L}$ is equal to $p_u$ and $\underset{i \in \mathscr{L}}{\mathds{P}}(w_m \leq \beta) = p_e$. For example, if one set $p_u=p_e=0.25$, the two parameters would equal the third and the first quartiles.
    \item For each layer $i$, find two local quantiles $\alpha_i, \beta_i \in [0, 1]$ such that $\underset{i}{\mathds{P}}(w_m \geq \alpha) = p_u$ where the probability is over only the $i$-th layer and $\underset{i}{\mathds{P}}(w_m \leq \beta) = p_e$. Figure \ref{fig:adaptive-router} illustrates how the quantiles are chosen.
    \item Decide the layer policy as in Table \ref{tab:layer-policy}. $\alpha_i > \alpha$ means the layer has relatively many uneven distributions, and $\beta_i < \beta$ means that the layer has relatively many even distributions. If $\alpha_i > \alpha$ and $\beta_i > \beta$ hold, then top-1 would be enough for the $i$-th layer, which justifies the static top-1 layer policy, and so on. For top-$k$ with $k \in \{ 1, 2, 3\}$, the layer's MoE block statically activates top $k$ experts. If both $\alpha_i > \alpha$ and $\beta_i < \beta$ hold, which implies that both extremely even and uneven distributions can occur within the layer, then the value of $k$ is decided token-wise; top-1 if $w_m \geq \alpha_i$, top-3 if $w_m \leq \beta_i$, and top-2 otherwise.
\end{enumerate}

\begin{table}
    \centering
    \begin{tabular}{c|c|c} \toprule
        {} & {$\alpha_i < \alpha$} & {$\alpha_i > \alpha$} \\ \hline
        $\beta_i > \beta$  & Top-2 & Top-1 \\ \hline
        $\beta_i < \beta$  & Top-3  & Dynamic (token-wise)  \\ \bottomrule
    \end{tabular}
    \caption{Layer-wise top-$k$ policy decision}
    \label{tab:layer-policy}
\end{table}

Here, token-wise dynamic top-$k$ is chosen in the only case $\alpha_i > \alpha$ and $\beta_i < \beta$, while other adaptive router approaches mainly use it. This is due to the observation that the static top-$k$, which only calls the top-$k$ function, has less latency than the dynamic one, which not only calls the top-$k$ function but also has to compare the maximal routing weight with $\alpha_i, \beta_i$.


\end{appendices}


\bibliographystyle{elsarticle-num-names}\biboptions{authoryear}
\bibliography{reference}

\begin{thebibliography}{45}
\expandafter\ifx\csname natexlab\endcsname\relax\def\natexlab#1{#1}\fi
\providecommand{\url}[1]{\texttt{#1}}
\providecommand{\href}[2]{#2}
\providecommand{\path}[1]{#1}
\providecommand{\DOIprefix}{doi:}
\providecommand{\ArXivprefix}{arXiv:}
\providecommand{\URLprefix}{URL: }
\providecommand{\Pubmedprefix}{pmid:}
\providecommand{\doi}[1]{\href{http://dx.doi.org/#1}{\path{#1}}}
\providecommand{\Pubmed}[1]{\href{pmid:#1}{\path{#1}}}
\providecommand{\bibinfo}[2]{#2}
\ifx\xfnm\relax \def\xfnm[#1]{\unskip,\space#1}\fi
\bibitem[{Cai et~al.(2024)Cai, Jiang, Wang, Tang, Kim, and Huang}]{survey1}
\bibinfo{author}{W.~Cai}, \bibinfo{author}{J.~Jiang}, \bibinfo{author}{F.~Wang}, \bibinfo{author}{J.~Tang}, \bibinfo{author}{S.~Kim}, \bibinfo{author}{J.~Huang},
\newblock \bibinfo{title}{A survey on mixture of experts},
\newblock \bibinfo{journal}{arXiv preprint arXiv:2407.06204}  (\bibinfo{year}{2024}).
\bibitem[{Shazeer et~al.(2017)Shazeer, Mirhoseini, Maziarz, Davis, Le, Hinton, and Dean}]{moe}
\bibinfo{author}{N.~Shazeer}, \bibinfo{author}{A.~Mirhoseini}, \bibinfo{author}{K.~Maziarz}, \bibinfo{author}{A.~Davis}, \bibinfo{author}{Q.~Le}, \bibinfo{author}{G.~Hinton}, \bibinfo{author}{J.~Dean},
\newblock \bibinfo{title}{Outrageously large neural networks: The sparsely-gated mixture-of-experts layer},
\newblock \bibinfo{journal}{arXiv preprint arXiv:1701.06538}  (\bibinfo{year}{2017}).
\bibitem[{Jiang et~al.(2024)Jiang, Sablayrolles, Roux, Mensch, Savary, Bamford, Chaplot, Casas, Hanna, Bressand et~al.}]{mixtral}
\bibinfo{author}{A.~Q. Jiang}, \bibinfo{author}{A.~Sablayrolles}, \bibinfo{author}{A.~Roux}, \bibinfo{author}{A.~Mensch}, \bibinfo{author}{B.~Savary}, \bibinfo{author}{C.~Bamford}, \bibinfo{author}{D.~S. Chaplot}, \bibinfo{author}{D.~d.~l. Casas}, \bibinfo{author}{E.~B. Hanna}, \bibinfo{author}{F.~Bressand}, et~al.,
\newblock \bibinfo{title}{Mixtral of experts},
\newblock \bibinfo{journal}{arXiv preprint arXiv:2401.04088}  (\bibinfo{year}{2024}).
\bibitem[{Zhang et~al.(2022)Zhang, Lin, Liu, Sun, and Zhou}]{moefication}
\bibinfo{author}{Z.~Zhang}, \bibinfo{author}{Y.~Lin}, \bibinfo{author}{Z.~Liu}, \bibinfo{author}{M.~Sun}, \bibinfo{author}{J.~Zhou},
\newblock \bibinfo{title}{Moefication: Transformer feed-forward layers are mixtures of experts},
\newblock \bibinfo{journal}{Findings of the Association for Computational Linguistics 2022}  (\bibinfo{year}{2022}) \bibinfo{pages}{877--890}.
\bibitem[{Raffel et~al.(2020)Raffel, Shazeer, Roberts, Lee, Narang, Matena, Zhou, Li, and Liu}]{t5}
\bibinfo{author}{C.~Raffel}, \bibinfo{author}{N.~Shazeer}, \bibinfo{author}{A.~Roberts}, \bibinfo{author}{K.~Lee}, \bibinfo{author}{S.~Narang}, \bibinfo{author}{M.~Matena}, \bibinfo{author}{Y.~Zhou}, \bibinfo{author}{W.~Li}, \bibinfo{author}{P.~J. Liu},
\newblock \bibinfo{title}{Exploring the limits of transfer learning with a unified text-to-text transformer},
\newblock \bibinfo{journal}{Journal of machine learning research} \bibinfo{volume}{21} (\bibinfo{year}{2020}) \bibinfo{pages}{1--67}.
\bibitem[{Devlin et~al.(2018)Devlin, Chang, Lee, and Toutanova}]{bert}
\bibinfo{author}{J.~Devlin}, \bibinfo{author}{M.-W. Chang}, \bibinfo{author}{K.~Lee}, \bibinfo{author}{K.~Toutanova},
\newblock \bibinfo{title}{Bert: Pre-training of deep bidirectional transformers for language understanding},
\newblock \bibinfo{journal}{arXiv preprint arXiv:1810.04805}  (\bibinfo{year}{2018}).
\bibitem[{Shazeer(2020)}]{swiglu}
\bibinfo{author}{N.~Shazeer},
\newblock \bibinfo{title}{Glu variants improve transformer},
\newblock \bibinfo{journal}{arXiv preprint arXiv:2002.05202}  (\bibinfo{year}{2020}).
\bibitem[{Zhu et~al.(2024)Zhu, Qu, Dong, Ruan, Tong, He, and Cheng}]{llama-moe}
\bibinfo{author}{T.~Zhu}, \bibinfo{author}{X.~Qu}, \bibinfo{author}{D.~Dong}, \bibinfo{author}{J.~Ruan}, \bibinfo{author}{J.~Tong}, \bibinfo{author}{C.~He}, \bibinfo{author}{Y.~Cheng},
\newblock \bibinfo{title}{Llama-moe: Building mixture-of-experts from llama with continual pre-training},
\newblock \bibinfo{journal}{arXiv preprint arXiv:2406.16554}  (\bibinfo{year}{2024}).
\bibitem[{Touvron et~al.(2023)Touvron, Lavril, Izacard, Martinet, Lachaux, Lacroix, Rozi{\`e}re, Goyal, Hambro, Azhar et~al.}]{llama}
\bibinfo{author}{H.~Touvron}, \bibinfo{author}{T.~Lavril}, \bibinfo{author}{G.~Izacard}, \bibinfo{author}{X.~Martinet}, \bibinfo{author}{M.-A. Lachaux}, \bibinfo{author}{T.~Lacroix}, \bibinfo{author}{B.~Rozi{\`e}re}, \bibinfo{author}{N.~Goyal}, \bibinfo{author}{E.~Hambro}, \bibinfo{author}{F.~Azhar}, et~al.,
\newblock \bibinfo{title}{Llama: Open and efficient foundation language models},
\newblock \bibinfo{journal}{arXiv preprint arXiv:2302.13971}  (\bibinfo{year}{2023}).
\bibitem[{Wu et~al.(2024)Wu, Qiu, Wang, Zhao, and Fu}]{wu2024}
\bibinfo{author}{H.~Wu}, \bibinfo{author}{Z.~Qiu}, \bibinfo{author}{Z.~Wang}, \bibinfo{author}{H.~Zhao}, \bibinfo{author}{J.~Fu},
\newblock \bibinfo{title}{Gw-moe: Resolving uncertainty in moe router with global workspace theory},
\newblock \bibinfo{journal}{arXiv preprint arXiv:2406.12375}  (\bibinfo{year}{2024}).
\bibitem[{Huang et~al.(2024)Huang, An, Zhuang, Tao, Zhang, Jin, Xu, Xu, Chen, Huang, and Feng}]{huang2024}
\bibinfo{author}{Q.~Huang}, \bibinfo{author}{Z.~An}, \bibinfo{author}{N.~Zhuang}, \bibinfo{author}{M.~Tao}, \bibinfo{author}{C.~Zhang}, \bibinfo{author}{Y.~Jin}, \bibinfo{author}{K.~Xu}, \bibinfo{author}{K.~Xu}, \bibinfo{author}{L.~Chen}, \bibinfo{author}{S.~Huang}, \bibinfo{author}{Y.~Feng},
\newblock \bibinfo{title}{Harder tasks need more experts: Dynamic routing in moe models},
\newblock \bibinfo{journal}{arXiv preprint arXiv:2403.07652}  (\bibinfo{year}{2024}).
\bibitem[{Li et~al.(2023)Li, Su, Yang, Jiang, Wang, and Xu}]{li2023}
\bibinfo{author}{J.~Li}, \bibinfo{author}{Q.~Su}, \bibinfo{author}{Y.~Yang}, \bibinfo{author}{Y.~Jiang}, \bibinfo{author}{C.~Wang}, \bibinfo{author}{H.~Xu},
\newblock \bibinfo{title}{Adaptive gating in mixture-of-experts based language models},
\newblock \bibinfo{journal}{arXiv preprint arXiv:2310.07188}  (\bibinfo{year}{2023}).
\bibitem[{Zeng et~al.(2024)Zeng, Miao, Gao, Zhang, and Deng}]{zeng2024}
\bibinfo{author}{Z.~Zeng}, \bibinfo{author}{Y.~Miao}, \bibinfo{author}{H.~Gao}, \bibinfo{author}{H.~Zhang}, \bibinfo{author}{Z.~Deng},
\newblock \bibinfo{title}{Adamoe: Token-adaptive routing with null experts for mixture-of-experts language models},
\newblock \bibinfo{journal}{arXiv preprint arXiv:2406.13233}  (\bibinfo{year}{2024}).
\bibitem[{Lu et~al.(2024)Lu, Liu, Xu, Zhou, Huang, Zhang, Yan, and Li}]{lu2024}
\bibinfo{author}{X.~Lu}, \bibinfo{author}{Q.~Liu}, \bibinfo{author}{Y.~Xu}, \bibinfo{author}{A.~Zhou}, \bibinfo{author}{S.~Huang}, \bibinfo{author}{B.~Zhang}, \bibinfo{author}{J.~Yan}, \bibinfo{author}{H.~Li},
\newblock \bibinfo{title}{Not all experts are equal: Efficient expert pruning and skipping for mixture-of-experts large language models},
\newblock \bibinfo{journal}{arXiv preprint arXiv:2402.14800}  (\bibinfo{year}{2024}).
\bibitem[{Hinton et~al.(2015)Hinton, Vinyals, and Dean}]{kd}
\bibinfo{author}{G.~Hinton}, \bibinfo{author}{O.~Vinyals}, \bibinfo{author}{J.~Dean},
\newblock \bibinfo{title}{Distilling the knowledge in a neural network},
\newblock \bibinfo{journal}{arXiv preprint arXiv:1503.02531}  (\bibinfo{year}{2015}).
\bibitem[{Jacobs et~al.(1991)Jacobs, Jordan, Nowlan, and Hinton}]{firstmoe}
\bibinfo{author}{R.~A. Jacobs}, \bibinfo{author}{M.~I. Jordan}, \bibinfo{author}{S.~J. Nowlan}, \bibinfo{author}{G.~E. Hinton},
\newblock \bibinfo{title}{Adaptive mixtures of local experts},
\newblock \bibinfo{journal}{Neural computation} \bibinfo{volume}{3} (\bibinfo{year}{1991}) \bibinfo{pages}{79--87}.
\bibitem[{Lepikhin et~al.(2020)Lepikhin, Lee, Xu, Chen, Firat, Huang, Krikun, Shazeer, and Chen}]{gshard}
\bibinfo{author}{D.~Lepikhin}, \bibinfo{author}{H.~Lee}, \bibinfo{author}{Y.~Xu}, \bibinfo{author}{D.~Chen}, \bibinfo{author}{O.~Firat}, \bibinfo{author}{Y.~Huang}, \bibinfo{author}{M.~Krikun}, \bibinfo{author}{N.~Shazeer}, \bibinfo{author}{Z.~Chen},
\newblock \bibinfo{title}{Gshard: Scaling giant models with conditional computation and automatic sharding},
\newblock \bibinfo{journal}{arXiv preprint arXiv:2006.16668}  (\bibinfo{year}{2020}).
\bibitem[{Fedus et~al.(2022)Fedus, Zoph, and Shazeer}]{switch}
\bibinfo{author}{W.~Fedus}, \bibinfo{author}{B.~Zoph}, \bibinfo{author}{N.~Shazeer},
\newblock \bibinfo{title}{Switch transformers: Scaling to trillion parameter models with simple and efficient sparsity},
\newblock \bibinfo{journal}{The Journal of Machine Learning Research} \bibinfo{volume}{23} (\bibinfo{year}{2022}) \bibinfo{pages}{5232--5270}.
\bibitem[{DeepSeek-AI(2024)}]{deepseek2}
\bibinfo{author}{DeepSeek-AI},
\newblock \bibinfo{title}{Deepseek-v2: A strong, economical, and efficient mixture-of-experts language model},
\newblock \bibinfo{journal}{arXiv preprint arXiv:2405.04434}  (\bibinfo{year}{2024}).
\bibitem[{Databricks(2024)}]{dbrx}
\bibinfo{author}{Databricks}, \bibinfo{title}{Introducing dbrx: A new state-of-the-art open llm}, \bibinfo{year}{2024}. \URLprefix \url{https://www.databricks.com/blog/ introducing-dbrx-new-state-art-open-llm}.
\bibitem[{xAI(2024)}]{grok}
\bibinfo{author}{xAI}, \bibinfo{title}{Grok-1}, \bibinfo{year}{2024}. \URLprefix \url{https://github.com/xai-org/grok-1}.
\bibitem[{Wei et~al.(2024)Wei, Zhu, Zhao, Cheng, Li, Lü, Cheng, Zhang, Zhang, Zeng, Wang, Ma, Hu, Yan, Fang, and Zhou}]{skywork-moe}
\bibinfo{author}{T.~Wei}, \bibinfo{author}{B.~Zhu}, \bibinfo{author}{L.~Zhao}, \bibinfo{author}{C.~Cheng}, \bibinfo{author}{B.~Li}, \bibinfo{author}{W.~Lü}, \bibinfo{author}{P.~Cheng}, \bibinfo{author}{J.~Zhang}, \bibinfo{author}{X.~Zhang}, \bibinfo{author}{L.~Zeng}, \bibinfo{author}{X.~Wang}, \bibinfo{author}{Y.~Ma}, \bibinfo{author}{R.~Hu}, \bibinfo{author}{S.~Yan}, \bibinfo{author}{H.~Fang}, \bibinfo{author}{Y.~Zhou},
\newblock \bibinfo{title}{Skywork-moe: A deep dive into training techniques for mixture-of-experts language models},
\newblock \bibinfo{journal}{arXiv preprint arXiv:2406.06563}  (\bibinfo{year}{2024}).
\bibitem[{Romero et~al.(2014)Romero, Ballas, Kahou, Chassang, Gatta, and Bengio}]{romero2014}
\bibinfo{author}{A.~Romero}, \bibinfo{author}{N.~Ballas}, \bibinfo{author}{S.~E. Kahou}, \bibinfo{author}{A.~Chassang}, \bibinfo{author}{C.~Gatta}, \bibinfo{author}{Y.~Bengio},
\newblock \bibinfo{title}{Fitnets: Hints for thin deep nets},
\newblock \bibinfo{journal}{arXiv preprint arXiv:1412.6550}  (\bibinfo{year}{2014}).
\bibitem[{Sun et~al.(2019)Sun, Cheng, Gan, and Liu}]{sun2019}
\bibinfo{author}{S.~Sun}, \bibinfo{author}{Y.~Cheng}, \bibinfo{author}{Z.~Gan}, \bibinfo{author}{J.~Liu},
\newblock \bibinfo{title}{Patient knowledge distillation for bert model compression},
\newblock \bibinfo{journal}{Proceedings of the 2019 Conference on Empirical Methods in Natural Language Processing and the 9th International Joint Conference on Natural Language Processing (EMNLP-IJCNLP)}  (\bibinfo{year}{2019}) \bibinfo{pages}{4323--4332}.
\bibitem[{Jiao et~al.(2020)Jiao, Yin, Shang, Jiang, Chen, Li, Wang, and Liu}]{tiny-bert}
\bibinfo{author}{X.~Jiao}, \bibinfo{author}{Y.~Yin}, \bibinfo{author}{L.~Shang}, \bibinfo{author}{X.~Jiang}, \bibinfo{author}{X.~Chen}, \bibinfo{author}{L.~Li}, \bibinfo{author}{F.~Wang}, \bibinfo{author}{Q.~Liu},
\newblock \bibinfo{title}{Tinybert: Distilling bert for natural language understanding},
\newblock \bibinfo{journal}{Findings of the Association for Computational Linguistics: EMNLP 2020}  (\bibinfo{year}{2020}) \bibinfo{pages}{4163--4174}.
\bibitem[{Liang et~al.(2023)Liang, Zuo, Zhang, He, Chen, and Zhao}]{chen2023}
\bibinfo{author}{C.~Liang}, \bibinfo{author}{S.~Zuo}, \bibinfo{author}{Q.~Zhang}, \bibinfo{author}{P.~He}, \bibinfo{author}{W.~Chen}, \bibinfo{author}{T.~Zhao},
\newblock \bibinfo{title}{Less is more: Task-aware layer-wise distillation for language model compression},
\newblock \bibinfo{journal}{ICML'23: Proceedings of the 40th International Conference on Machine Learning}  (\bibinfo{year}{2023}) \bibinfo{pages}{20852--20867}.
\bibitem[{Zhang et~al.(2022)Zhang, Shen, Huang, Zhou, Rong, and Xiong}]{zhang2022}
\bibinfo{author}{X.~Zhang}, \bibinfo{author}{Y.~Shen}, \bibinfo{author}{Z.~Huang}, \bibinfo{author}{J.~Zhou}, \bibinfo{author}{W.~Rong}, \bibinfo{author}{Z.~Xiong},
\newblock \bibinfo{title}{Mixture of attention heads: Selecting attention heads per token},
\newblock \bibinfo{journal}{arXiv preprint arXiv:2210.05144}  (\bibinfo{year}{2022}).
\bibitem[{Shen et~al.(2024)Shen, Guo, Cai, and Qin}]{jetmoe}
\bibinfo{author}{Y.~Shen}, \bibinfo{author}{Z.~Guo}, \bibinfo{author}{T.~Cai}, \bibinfo{author}{Z.~Qin},
\newblock \bibinfo{title}{Jetmoe: Reaching llama2 performance with 0.1 m dollars},
\newblock \bibinfo{journal}{arXiv preprint arXiv:2404.07413}  (\bibinfo{year}{2024}).
\bibitem[{Li et~al.(2022)Li, You, Bhojanapalli, Li, Rawat, Reddi, Ye, Chern, Yu, Guo et~al.}]{li2022}
\bibinfo{author}{Z.~Li}, \bibinfo{author}{C.~You}, \bibinfo{author}{S.~Bhojanapalli}, \bibinfo{author}{D.~Li}, \bibinfo{author}{A.~S. Rawat}, \bibinfo{author}{S.~J. Reddi}, \bibinfo{author}{K.~Ye}, \bibinfo{author}{F.~Chern}, \bibinfo{author}{F.~Yu}, \bibinfo{author}{R.~Guo}, et~al.,
\newblock \bibinfo{title}{The lazy neuron phenomenon: On emergence of activation sparsity in transformers},
\newblock \bibinfo{journal}{arXiv preprint arXiv:2210.06313}  (\bibinfo{year}{2022}).
\bibitem[{Liu et~al.(2023)Liu, Wang, Dao, Zhou, Yuan, Song, Shrivastava, Zhang, Tian, Re et~al.}]{liu2023}
\bibinfo{author}{Z.~Liu}, \bibinfo{author}{J.~Wang}, \bibinfo{author}{T.~Dao}, \bibinfo{author}{T.~Zhou}, \bibinfo{author}{B.~Yuan}, \bibinfo{author}{Z.~Song}, \bibinfo{author}{A.~Shrivastava}, \bibinfo{author}{C.~Zhang}, \bibinfo{author}{Y.~Tian}, \bibinfo{author}{C.~Re}, et~al.,
\newblock \bibinfo{title}{Deja vu: Contextual sparsity for efficient llms at inference time},
\newblock in: \bibinfo{booktitle}{International Conference on Machine Learning}, \bibinfo{organization}{PMLR}, \bibinfo{year}{2023}, pp. \bibinfo{pages}{22137--22176}.
\bibitem[{Zheng et~al.(2024)Zheng, Bai, Chen, Lai, and Prakash}]{zheng2024}
\bibinfo{author}{H.~Zheng}, \bibinfo{author}{X.~Bai}, \bibinfo{author}{B.~Chen}, \bibinfo{author}{F.~Lai}, \bibinfo{author}{A.~Prakash},
\newblock \bibinfo{title}{Learn to be efficient: Build structured sparsity in large language models},
\newblock \bibinfo{journal}{arXiv preprint arXiv:2402.06126}  (\bibinfo{year}{2024}).
\bibitem[{Pan et~al.(2024)Pan, Shen, Liu, Mishra, Zhang, Oliva, Raffel, and Panda}]{pan2024}
\bibinfo{author}{B.~Pan}, \bibinfo{author}{Y.~Shen}, \bibinfo{author}{H.~Liu}, \bibinfo{author}{M.~Mishra}, \bibinfo{author}{G.~Zhang}, \bibinfo{author}{A.~Oliva}, \bibinfo{author}{C.~Raffel}, \bibinfo{author}{R.~Panda},
\newblock \bibinfo{title}{Dense training, sparse inference: Rethinking training of mixture-of-experts language models},
\newblock \bibinfo{journal}{arXiv preprint arXiv:2404.05567}  (\bibinfo{year}{2024}).
\bibitem[{Yun et~al.(2024)Yun, Zhuang, Fu, Xing, and Zhang}]{yun2024}
\bibinfo{author}{L.~Yun}, \bibinfo{author}{Y.~Zhuang}, \bibinfo{author}{Y.~Fu}, \bibinfo{author}{E.~P. Xing}, \bibinfo{author}{H.~Zhang},
\newblock \bibinfo{title}{Toward inference-optimal mixture-of-expert large language models},
\newblock \bibinfo{journal}{arXiv preprint arXiv:2404.02852}  (\bibinfo{year}{2024}).
\bibitem[{Krajewski et~al.(2024)Krajewski, Ludziejewski, Adamczewski, Pi{\'o}ro, Krutul, Antoniak, Ciebiera, Kr{\'o}l, Odrzyg{\'o}{\'z}d{\'z}, Sankowski et~al.}]{krajewski2024}
\bibinfo{author}{J.~Krajewski}, \bibinfo{author}{J.~Ludziejewski}, \bibinfo{author}{K.~Adamczewski}, \bibinfo{author}{M.~Pi{\'o}ro}, \bibinfo{author}{M.~Krutul}, \bibinfo{author}{S.~Antoniak}, \bibinfo{author}{K.~Ciebiera}, \bibinfo{author}{K.~Kr{\'o}l}, \bibinfo{author}{T.~Odrzyg{\'o}{\'z}d{\'z}}, \bibinfo{author}{P.~Sankowski}, et~al.,
\newblock \bibinfo{title}{Scaling laws for fine-grained mixture of experts},
\newblock \bibinfo{journal}{arXiv preprint arXiv:2402.07871}  (\bibinfo{year}{2024}).
\bibitem[{Dai et~al.(2024)Dai, Deng, Zhao, Xu, Gao, Chen, Li, Zeng, Yu, Wu et~al.}]{deepseek}
\bibinfo{author}{D.~Dai}, \bibinfo{author}{C.~Deng}, \bibinfo{author}{C.~Zhao}, \bibinfo{author}{R.~Xu}, \bibinfo{author}{H.~Gao}, \bibinfo{author}{D.~Chen}, \bibinfo{author}{J.~Li}, \bibinfo{author}{W.~Zeng}, \bibinfo{author}{X.~Yu}, \bibinfo{author}{Y.~Wu}, et~al.,
\newblock \bibinfo{title}{Deepseekmoe: Towards ultimate expert specialization in mixture-of-experts language models},
\newblock \bibinfo{journal}{arXiv preprint arXiv:2401.06066}  (\bibinfo{year}{2024}).
\bibitem[{Zoph et~al.(2022)Zoph, Bello, Kumar, Du, Huang, Dean, Shazeer, and Fedus}]{stmoe}
\bibinfo{author}{B.~Zoph}, \bibinfo{author}{I.~Bello}, \bibinfo{author}{S.~Kumar}, \bibinfo{author}{N.~Du}, \bibinfo{author}{Y.~Huang}, \bibinfo{author}{J.~Dean}, \bibinfo{author}{N.~Shazeer}, \bibinfo{author}{W.~Fedus},
\newblock \bibinfo{title}{St-moe: Designing stable and transferable sparse expert models},
\newblock \bibinfo{journal}{arXiv preprint arXiv:2202.08906}  (\bibinfo{year}{2022}).
\bibitem[{Komatsuzaki et~al.(2022)Komatsuzaki, Puigcerver, Lee-Thorp, Riquelme~Ruiz, Mustafa, Ainslie, Tay, Dehghani, and Houlsby}]{sparse-upcycling}
\bibinfo{author}{A.~Komatsuzaki}, \bibinfo{author}{J.~Puigcerver}, \bibinfo{author}{J.~Lee-Thorp}, \bibinfo{author}{C.~Riquelme~Ruiz}, \bibinfo{author}{B.~Mustafa}, \bibinfo{author}{J.~Ainslie}, \bibinfo{author}{Y.~Tay}, \bibinfo{author}{M.~Dehghani}, \bibinfo{author}{N.~Houlsby},
\newblock \bibinfo{title}{Sparse upcycling: Training mixture-of-experts from dense checkpoints},
\newblock \bibinfo{journal}{arXiv preprint arXiv:2212.05055}  (\bibinfo{year}{2022}).
\bibitem[{Zuo et~al.(2022)Zuo, Zhang, Liang, He, Zhao, and Chen}]{moebert}
\bibinfo{author}{S.~Zuo}, \bibinfo{author}{Q.~Zhang}, \bibinfo{author}{C.~Liang}, \bibinfo{author}{P.~He}, \bibinfo{author}{T.~Zhao}, \bibinfo{author}{W.~Chen},
\newblock \bibinfo{title}{Moebert: from bert to mixture-of-experts via importance-guided adaptation},
\newblock \bibinfo{journal}{arXiv preprint arXiv:2204.07675}  (\bibinfo{year}{2022}).
\bibitem[{Kim et~al.(2023)Kim, Park, Kim, Lee, Song, Kim, Kim, Kim, Lee, Kim, Ahn, Yang, Lee, Park, Gim, Cha, Lee, and Kim}]{solar}
\bibinfo{author}{D.~Kim}, \bibinfo{author}{C.~Park}, \bibinfo{author}{S.~Kim}, \bibinfo{author}{W.~Lee}, \bibinfo{author}{W.~Song}, \bibinfo{author}{Y.~Kim}, \bibinfo{author}{H.~Kim}, \bibinfo{author}{Y.~Kim}, \bibinfo{author}{H.~Lee}, \bibinfo{author}{J.~Kim}, \bibinfo{author}{C.~Ahn}, \bibinfo{author}{S.~Yang}, \bibinfo{author}{S.~Lee}, \bibinfo{author}{H.~Park}, \bibinfo{author}{G.~Gim}, \bibinfo{author}{M.~Cha}, \bibinfo{author}{H.~Lee}, \bibinfo{author}{S.~Kim},
\newblock \bibinfo{title}{Solar 10.7b: Scaling large language models with simple yet effective depth up-scaling},
\newblock \bibinfo{journal}{arXiv preprint arXiv:2312.15166}  (\bibinfo{year}{2023}).
\bibitem[{Palla(2023)}]{chatbot}
\bibinfo{author}{A.~Palla}, \bibinfo{title}{Chatbot instruction prompts}, \bibinfo{year}{2023}. \URLprefix \url{https://huggingface.co/datasets/alespalla/chatbot\_instruction\_prompts}.
\bibitem[{Guo et~al.(2024)Guo, Cheng, Tang, and Lin}]{guo2024}
\bibinfo{author}{Y.~Guo}, \bibinfo{author}{Z.~Cheng}, \bibinfo{author}{X.~Tang}, \bibinfo{author}{T.~Lin},
\newblock \bibinfo{title}{Dynamic mixture of experts: An auto-tuning approach for efficient transformer models},
\newblock \bibinfo{journal}{arXiv preprint arXiv:2405.14297}  (\bibinfo{year}{2024}).
\bibitem[{Rasley et~al.(2020)Rasley, Rajbhandari, Ruwase, and He}]{deepspeed}
\bibinfo{author}{J.~Rasley}, \bibinfo{author}{S.~Rajbhandari}, \bibinfo{author}{O.~Ruwase}, \bibinfo{author}{Y.~He},
\newblock \bibinfo{title}{Deepspeed: System optimizations enable training deep learning models with over 100 billion parameters},
\newblock \bibinfo{journal}{Proceedings of the 26th ACM SIGKDD International Conference on Knowledge Discovery \& Data Mining}  (\bibinfo{year}{2020}) \bibinfo{pages}{3505--3506}.
\bibitem[{Kwon et~al.(2023)Kwon, Li, Zhuang, Sheng, Zheng, Yu, Gonzalez, Zhang, and Stoica}]{vllm}
\bibinfo{author}{W.~Kwon}, \bibinfo{author}{Z.~Li}, \bibinfo{author}{S.~Zhuang}, \bibinfo{author}{Y.~Sheng}, \bibinfo{author}{L.~Zheng}, \bibinfo{author}{C.~H. Yu}, \bibinfo{author}{J.~Gonzalez}, \bibinfo{author}{H.~Zhang}, \bibinfo{author}{I.~Stoica},
\newblock \bibinfo{title}{Efficient memory management for large language model serving with pagedattention},
\newblock \bibinfo{journal}{Proceedings of the 29th Symposium on Operating Systems Principles}  (\bibinfo{year}{2023}) \bibinfo{pages}{611--626}.
\bibitem[{Cobbe et~al.(2021)Cobbe, Kosaraju, Bavarian, Chen, Jun, Kaiser, Plappert, Tworek, Hilton, Nakano, Hesse, and Schulman}]{gsm8k}
\bibinfo{author}{K.~Cobbe}, \bibinfo{author}{V.~Kosaraju}, \bibinfo{author}{M.~Bavarian}, \bibinfo{author}{M.~Chen}, \bibinfo{author}{H.~Jun}, \bibinfo{author}{L.~Kaiser}, \bibinfo{author}{M.~Plappert}, \bibinfo{author}{J.~Tworek}, \bibinfo{author}{J.~Hilton}, \bibinfo{author}{R.~Nakano}, \bibinfo{author}{C.~Hesse}, \bibinfo{author}{J.~Schulman},
\newblock \bibinfo{title}{Training verifiers to solve math word problems},
\newblock \bibinfo{journal}{arXiv preprint arXiv:2110.14168}  (\bibinfo{year}{2021}).
\bibitem[{Gao et~al.(2020)Gao, Biderman, Black, Golding, Hoppe, Foster, Phang, He, Thite, Nabeshima et~al.}]{pile}
\bibinfo{author}{L.~Gao}, \bibinfo{author}{S.~Biderman}, \bibinfo{author}{S.~Black}, \bibinfo{author}{L.~Golding}, \bibinfo{author}{T.~Hoppe}, \bibinfo{author}{C.~Foster}, \bibinfo{author}{J.~Phang}, \bibinfo{author}{H.~He}, \bibinfo{author}{A.~Thite}, \bibinfo{author}{N.~Nabeshima}, et~al.,
\newblock \bibinfo{title}{The pile: An 800gb dataset of diverse text for language modeling},
\newblock \bibinfo{journal}{arXiv preprint arXiv:2101.00027}  (\bibinfo{year}{2020}).

\end{thebibliography}






\end{document}